# Deep learning ensembles for melanoma recognition in dermoscopy images[1]

N. C. F. Codella, Q. B. Nguyen, S. Pankanti, D. Gutman, B. Helba, A. Halpern, J. R. Smith

## Abstract

Melanoma is the deadliest form of skin cancer. While curable with early detection, only highly trained specialists are capable of accurately recognizing the disease. As expertise is in limited supply, automated systems capable of identifying disease could save lives, reduce unnecessary biopsies, and reduce costs. Toward this goal, we propose a system that combines recent developments in deep learning with established machine learning approaches, creating ensembles of methods that are capable of segmenting skin lesions, as well as analyzing the detected area and surrounding tissue for melanoma detection. The system is evaluated using the largest publicly available benchmark dataset of dermoscopic images, containing 900 training and 379 testing images. New state-of-the-art performance levels are demonstrated, leading to an improvement in the area under receiver operating characteristic curve of 7.5% (0.843 vs. 0.783), in average precision of 4% (0.649 vs. 0.624), and in specificity measured at the clinically relevant 95% sensitivity operating point 2.9 times higher than the previous state-of-the-art (36.8% specificity compared to 12.5%). Compared to the average of 8 expert dermatologists on a subset of 100 test images, the proposed system produces a higher accuracy (76% vs. 70.5%), and specificity (62% vs. 59%) evaluated at an equivalent sensitivity (82%).

## Introduction

Skin cancer is the most common cancer in the United States, with over 5 million cases diagnosed each year [1]. Melanoma, the deadliest form of skin cancer, is involved in approximately 100,000 new instances every year in the United States, and over 9,000 deaths [2]. The cost to the U.S. healthcare system exceeds $8 billion [3]. Internationally, skin cancer also poses a major public health threat. In Australia, there are over 13,000 new instances of melanoma yearly, leading to over 1,200 deaths [4]. In Europe, melanoma causes over 20,000 deaths a year [5].

In order to combat the rising mortality of melanoma, early detection is critical. Currently, highly trained experts and professional equipment are necessary for accurate and early detection of melanoma. Dermoscopy is a specialized method of high-resolution imaging of the skin that reduces skin surface reflectance, allowing clinicians to visualize deeper underlying structures. Using this device, specially trained clinicians have demonstrated a diagnostic accuracy as high as 75-84% [7]. However, recognition performance drops significantly when the clinicians are not adequately trained [8, 9].

While in the United States there are over 10,000 dermatologists, in other areas of the world the supply of expertise is limited. For example, in Australia, the number of registered dermatologists in 2004 was approximately 340 [10], and in New Zealand, there were 16 [11]. Restricted access to expert consultation leads to additional challenges in providing adequate levels of care to the populations that are at risk.

In order to address the limited supply of experts, there has been effort in the research community to develop automated image analysis systems to detect disease from dermoscopy images. Such technology could be used as a diagnostic tool by primary care physicians and staff for regular screening, or by clinicians who are otherwise not trained to interpret dermoscopy images. Review articles covering a spectrum of publications have been recently presented [7, 12-15]. The variety of automated image analysis techniques discussed is broad, but mostly restricted within the space of classical computer vision approaches, typically using combinations of low-level visual feature representations (color, edge, and texture descriptors, quantification of melanin based on color, etc.), rule-based image processing or segmentation algorithms, and classical machine learning techniques, such as *k*-nearest neighbor (kNN) and support vector machines (SVM). Some publications have presented algorithms that include segmentation of the lesion [16-21]. A team from the Pedro Hispano Hospital of Portugal sought to evaluate the performance of several (e.g., SVM and kNN) machine learning classifiers based on color, edge, and texture descriptors [22,23]. Other teams employed ensemble learning approaches [24-26]. Interestingly, some earlier work employed neural network machine learning approaches [27-31]. However, these were built on top of hand-coded low-level features.

---

[1] This paper will appear, in final form, in the *IBM Journal of Research and Development*, vol. 61, no. 4/5, 2017, as part of a special issue on "Deep Learning." Please cite the IBM Journal official paper version of record. For more information on the journal, see: *http://www.research.ibm.com/journal/*. ©IBM.

More recent work has begun to examine the efficacy of the state-of-the-art deep learning approaches to image recognition within the dermatology and dermoscopy application domain [32,33]. Representations learned from the natural photo domain were leveraged, in conjunction with unsupervised and hand-coded features, to achieve state-of-the-art performance in a data of over 2,000 dermoscopy images [32]. However, the work was limited to lesion images that had been manually pre-segmented: images were already cropped around the lesion of interest.

In 2016, the International Skin Imaging Collaboration (ISIC) organized an international effort to aggregate a dataset of dermoscopic images from multiple institutions for the purposes of developing and evaluating clinical and automated techniques for the diagnosis of melanoma [34]. A snapshot of the dataset that contained the most complete set of annotations was selected to host a melanoma recognition challenge at the 2016 International Symposium on Biomedical Imaging (ISBI 2016). The challenge was titled "Skin Lesion Analysis toward Melanoma Detection" [35]. In total, 38 individual participants contributed 79 submissions across 3 image analysis tasks, including 43 submissions toward disease classification. This was the first publicly organized large-scale standardized evaluation of algorithms for the detection of melanoma. Top performing techniques involved deep learning approaches, including Deep Residual Networks for classification [36], and fully convolutional networks for segmentation [37,38].

In this work, we combine hand-coded feature extractors, sparse-coding methods, and SVMs, with more recent machine learning techniques, including deep residual networks and fully convolutional neural networks, into ensembles focused toward the task of melanoma recognition and segmentation in dermoscopy images. We have chosen to use the ISBI 2016 dataset for evaluation, which provides an immediate comparison to dozens of prior algorithms, and opportunity for future comparisons. New state-of-the-art performance levels are demonstrated across a variety of evaluation metrics, including an almost tripling of specificity measured at 95% sensitivity. These results emphasize that combining a multitude of machine learning approaches can yield higher performance than relying on any one method alone, especially in regards to recognition of melanoma in dermoscopic images.

## Dataset

For the training and evaluation, we used the dataset released by the International Skin Imaging Collaboration (ISIC) for the 2016 International Symposium on Biomedical Imaging (ISBI 2016) challenge titled "Skin Lesion Analysis toward Melanoma Detection" [35]. The ISBI 2016 challenge dataset contains 900 annotated dermoscopic images for training (173 melanomas), and 379 images in a held-out test set for evaluation (75 melanomas). **Figure 1** shows sample images from this dataset. The challenge consisted of three parts: Part 1) Lesion Segmentation, Part 2) Lesion Dermoscopic Feature Extraction, and Part 3) Lesion Classification. *Parts 2 and 3* were further broken into *Parts 2 and 2B*, and *Parts 3 and 3B*. Part 2 represented the dermoscopic feature extraction problem as a classification task, whereas 2B represented the problem as a segmentation task. The purpose of this breakdown was to quantify how the framing of the problem influenced end system performance. Part 3 presented the disease classification task withholding ground truth (GT) segmentations from the held-out test dataset, whereas Part 3B provided the GT lesion segmentations with the held-out test dataset. The purpose of this breakdown was to understand how knowledge of ground truth segmentation affects disease classification performance.

For lesion segmentation tasks, outputs were represented as binary image masks. Pixels inside the lesions were represented by the pixel intensity value of 255, and pixels outside the lesions (background normal skin) were represented by pixel intensity values of 0. Participants were ranked according to the Jaccard index (JACC) [35]. Other segmentation performance metrics such as pixel-wise accuracy (ACC), sensitivity (SENS), and specificity (SPEC) were also reported. For lesion classification tasks, participants were ranked according to the average precision (AP). Additional performance metrics such as classification accuracy (ACC), sensitivity (SENS), specificity (SPEC), area under the receiver operator curve (AUC), and specificity measured at the clinically relevant 95% sensitivity threshold (SP95), were reported as well. The last metric is particularly important, as clinicians tend to desire systems that are set at a sensitivity level not likely to miss instances of disease. A detailed description of the definition of each metric is given in the ISBI challenge report [35].

The scope of this paper encompasses Parts 1 and 3 of the ISBI challenge. Part 2, and the potential influence of dermoscopic feature extraction towards the detection of disease, is left for a future study.

**Visual Recognition System**

The proposed visual recognition system consists of two primary components: segmentation and classification. With segmentation, the skin lesion is identified in the dermoscopic image and distinguished from background healthy skin. This allows the system to subsequently perform analysis within two contexts. The first context is focused within the lesion itself, the potentially diseased tissue, and the second context is focused within the entire image, including the surrounding area that may exhibit other patterns indicative of the disease state of the lesion.

*Segmentation*

For lesion segmentation, a fully convolutional network structure [38], similar to that used for the U-Net architecture [37], was implemented using Theano, Lasagne, and Nolearn python packages (master branch versions as of March 15, 2016). Generally speaking, the approach is a modeling framework that learns a functional mapping from an input image to an output image. The input image is the original image, and the output image is a segmentation mask. The structure of the network involves a series of convolution and pooling operations, followed by a single fully connected layer, and followed-up with a series of unpooling and deconvolution operations. Skip connections are used to link convolutional data prior to pooling operations with the deconvolution operations. This enables the network to model functional residuals, as well to supply higher resolution information to the output layers, in order to improve performance of the network in comparison to networks without the skip connections.

The network structure is depicted in **Figure 2**. In the proposed model, three stages of convolutions (abbreviated as "conv" in figures and tables) and pooling operations are followed by a fully-connected layer (abbreviated as "fc" in figures and tables), whereupon the process is symmetrically reversed with 3 stages of unpooling and deconvolution layers. In each stage prior to the fully connected layer, there are 3 convolution operations, followed by 1 pooling operation. The number of convolution filters for each convolution layer is doubled across stages, so that the second stage has twice as many filters as the first stage, and the third stage has 4 times as many filters as the first stage. After the fully connected layer, the ordering of the layers is simply reversed to 1 unpooling layer followed by 3 deconvolutional layers. Skip-connections (also referred to as concatenation layers) directly link the output of the last convolutional layer of stages prior to the fully connected layer, to the corresponding unpooling layer in the symmetrical stage on the opposite side of the fully connected layer. Across each stage after the fully connected layer, the number of convolution filters is halved, so that the last stage has as many filters again as the first stage before the fully connected layer.

The hyperbolic tangent was selected as the nonlinear squashing function for all neuron outputs. Gaussian noise layers were inserted after the input layer, after each pooling layer prior to the fully connected layer, and to the fully connected layer itself. Dropout was applied to all pooling layers before the fully connected layer, to the fully connected layer itself, and to all concatenation layers.

Images were input into the network using six color channels, including Red-Green-Blue (RGB) and Hue-Saturation-Value (HSV) color spaces. Empirical experimentation found improvements in performance on training data using 6 color channels as opposed to 3. All images were resized to 128-by-128 dimensions using bilinear interpolation. (Although the size of the input can be changed, for reduction of problem complexity, we held this value constant.) Input images are standard normalized by subtracting the mean pixel intensity and dividing the standard deviation for each image and each color channel independently. Output masks were normalized by dividing by the maximum pixel value (255), subtracting by 0.5, and multiplying by 1.9. This caused pixels in the mask representing background skin to take on values near -1, and pixels in the mask representing lesion to take on values near 1 (to convert network outputs back to image masks, this range is simply rescaled back to the range between 0 and 255). Output images were kept the same size as the input images.

Images input to the network were subjected to data augmentation within each training batch. Images are rotated, flipped, rescaled, shifted, cropped, and further subjected to non-linear distortions (sinusoidal remapping with varying phase, frequency, and amplitude, in both *X* and *Y* directions) [37]. The motivation for non-linear distortion is that it appropriately models the variation of soft-tissue biological structures.

The network has several tunable parameters. The first is the size of the convolution kernels, which is held constant across all stages before the fully connected layer, but is doubled and incremented in the first convolutional layers of stages on the opposite side of the fully connected layer. The second is the number of convolution filters in the convolutional layers of the first stage. The third parameter is the standard deviation of the noise in each Gaussian noise layer. The fourth is the degree of dropout used both prior to the fully connected layer, within the fully connected layer, and after the fully connected layer. The last two parameters are the learning rate and the momentum.

Training data for segmentation was split into two partitions of 80% (720) for training and 20% (180) for validation. Network parameters were trained on the 80%, and the loss function was computed on the 20%. Early stopping was employed if performance on the validation dataset did not improve within 100 epochs, if training ceased, and if the set of network weights that produced best performance on the validation split were saved to disk. In addition, the learning rate and momentum parameters were adapted as follows during the training phase: both were linearly adjusted from their initial starting values, which may vary based on the input, to 0.001 and 0.99, respectively, from the first epoch to the maximum epoch. This led to a gradual decrease in the learning rate and a gradual increase in momentum over the successive epochs. The network was trained as a regression problem using a squared-error loss function. The output of the network therefore represents a "soft" confidence spectrum rather than a hard binary decision, which helps the network model the segmentation function in the presence of ambiguity and human variability in ground truth annotations.

Two sets of experiments were conducted to train and assess the performance of the fully convolutional U-Net architecture for semantic segmentation of dermoscopic lesions. In the first experiment, a grid-search optimization process was employed to examine the configuration of tunable network parameters to minimize the final loss function value on the 20% validation dataset. The network that yielded the best performance on the validation split was applied to the test set to perform final segmentation. In the second, an ensemble of 10 networks, each with different parameter values, was created. The outputs of the 10 were averaged to create a combined segmentation prediction.

## *Classification*

For disease classification, we employ an ensemble of recent machine learning methods, including deep residual networks [36], convolutional neural networks [39], and fully convolutional U-Net architecture [37], as well as the established machine learning procedures, such as sparse coding, and hand-coded feature representations. Our method is an improvement over the previous methods used in the literature (e.g., [32]) in terms of use of automated segmentation, multi-contextual analysis, and additional machine learning techniques. Each machine learning technique was used to extract information in the form of a feature vector from the dermoscopic image in up to two contexts: that from the entire image, and that from a region tightly cropped around the lesion segmented by the segmentation procedure. Other prior reports have found this technique of considering more than one context to yield improvements in recognition performance [23,27]. Once the feature vectors have been extracted, a non-linear support vector machine (SVM) was used to learn a classifier over the feature vector to discriminate for melanoma. A histogram intersection kernel was employed, along with sigmoid feature normalization. The output of the SVM was calibrated to a logistic function (using three-fold cross-validation), roughly approximating the probability of melanoma on an image, assuming a balanced prior [32, 40-42]. After score calibration, the output of the SVMs trained over each feature independently was averaged, producing a final disease confidence score between 0.0 and 1.0 (0% and 100%), with a default decision threshold of 0.5 (50%). A visual overview of the approach is shown in **Figure 3**. In the following subsections, we describe each method used to extract features from the dermoscopic images.

*Hand-coded Feature Extraction*

Low-level visual features used in this work included color histogram, edge histogram, and a multi-scale variant of color local binary patterns (LBP) [42,43]. All of these features have been used in systems presented in prior literature that have achieved top performance in various medical image datasets [42], including those of dermoscopic images [32].

The color histogram distribution represents a 166-dimensional histogram in HSV color space. The edge histogram contains 8 edge direction bins and 8 edge magnitude bins, based on a Sobel filter (64-dimensional). Multiscale color LBP is an extension of the common grayscale LBP, whereby LBP descriptors are extracted across 4 color channels (Red, Green, Blue, and Hue), with one histogram per color channel. For each color channel, LBP descriptors are extracted across multiple scales (1/1, 1/2, 1/4, and 1/8th image size), and aggregated into a single histogram, weighted by the inverse of the spatial scale. For a 59-bin LBP histogram, this results in $59 \times 4 = 236$ total bins. These features were extracted from the two contexts of whole image, and lesion cropped regions.

*Sparse Coding*

Sparse coding (SC) is a class of unsupervised methods that learns a dictionary of sparse codes from which a given dataset can be reconstructed. The SPArse Modeling Software (SPAMS) sparse coding dictionary learning algorithm [44] is an online optimization approach for this method, based on stochastic approximations. Because the algorithm is efficient and has been used in the prior state-of-the-art melanoma recognition systems [32], the SPAMS algorithm was employed to learn dictionaries on this dataset. Four dictionaries were constructed in RGB and grayscale color spaces, across both the whole image context and the lesion cropped region context. Images were resized to $128 \times 128$ pixel dimensions before extraction of $8 \times 8$ patches, to learn dictionaries of 1024 elements. We used default parameters ($\lambda = 0.15$, and the number of iterations = 1000) recommended in the SPAMS implementation for minimization of the objective function. These features were extracted from the two contexts of whole image, and lesion cropped regions.

*Convolutional Neural Network*

The Caffe convolutional neural network (CNN) architecture developed at Berkeley [45] was used to extract image descriptors, similar to prior work [32]. A pre-trained model from the Image Large Scale Visual Recognition Challenge (ILSVRC) 2012 is provided for download from the website. This pre-trained model includes 5 convolutional layers, 2 fully connected layers, and a final 1000-dimensional concept detector layer. In this work, the first fully connected layer (4096 dimensions, referred to as "FC6"), is used as a visual descriptor for dermoscopy images. These features were extracted from the two contexts of whole image, and lesion cropped regions.

*Deep Residual Network*

The Deep Residual Network (DRN) is the most recent network structure to win the ImageNet recognition challenge [36]. The network containing 101 layers was used to extract a 1000-dimensional concept detector vector from dermoscopy images at the context of the whole image level. Prior state-of-the-art medical image recognition works have found that concept detector feature vectors contribute complimentary information to classification systems, improving performance over baselines lacking the feature vectors [32,42]. This feature was extracted only from the whole image context.

*Fully Convolutional U-Net*

The fully convolutional U-Net architecture used in this work for lesion segmentation was also used in the classification framework as a shape descriptor. The fully connected layer, which maximally compresses information required to reconstruct the image segmentation, was extracted from the network for every image in the dataset to serve as this feature vector. We trained an independent U-Net network and limit the dimensionality of the fully connected layer to 1024 dimensions to maintain compactness of the representation. This feature was extracted only from the whole image context, as the network is designed to operate within this context.

## Experimental Results

### *Segmentation*

**Table 1** presents a summary of the segmentation experiments. For the first experiment using an optimized single U-Net structure, we used a convolution kernel size of 5×5, pooling layers of 2×2, 32 convolution filters at the first stage, an 8192-dimensional fully connected layer, a dropout of 0.5 at all dropout layers, Gaussian noise with standard deviation of 0.025 applied to all noise layers, an initial learning rate of 0.01, an initial momentum of 0.95, and a maximum number of epochs set to 2000. This set of parameters produced the minimal error (0.0962) on the validation dataset split during grid-search over the parameter space. On the 379 images in the held-out test data, this network produced a Jaccard index of 0.836, an accuracy of 94.9%, sensitivity of 91.4%, and specificity of 96.3%. This performance places it at the second rank in the 2016 ISBI challenge on melanoma detection, 0.007 points below the Jaccard index of the top performer. Without data augmentation techniques beyond standard rotations and flips, performance of the network dropped to a 0.828 Jaccard Index, 94.7% accuracy, 91.2% sensitivity, and 96% specificity. Additionally, eliminating Gaussian noise and dropout reduced performance further to a 0.812 Jaccard Index, 94.1% accuracy, 89.8% sensitivity, and 95.9% specificity.

In the second experiment, a U-Net Ensemble of networks, each with varied network parameters, was employed, rather than relying on a single optimized network alone. 10 independently trained networks were created. For each network, parameters were selected to achieve a reasonable spectrum across network topology and dropout parameter values: the major network topology and dropout parameters are configured in at least 2 different values. The exact parameters used across all 10 networks are shown in **Table 2**. The resulting predictions of all 10 networks were averaged before being subjected to binary thresholding at a pixel value of 128. The resulting Jaccard index was 0.841, with corresponding accuracy of 95.1%, sensitivity of 91.1%, and specificity of 96.7%. This performance still places the network at the second rank in the challenge, but behind by merely 0.002 points as opposed to 0.007, a reduction of 71.4% of the remaining error behind the top ranked submission. Example segmentations prior to thresholding, along with the original image and the ground truth segmentation, are shown in **Figure 4**.

The residual error between automated algorithms and the ground truth was at least in part related to segmentation variability in the ground truth itself, which results from intrinsic differences in human annotation, low contrast, presence of hair, or visual adjacency of the lesion to other lesions in the peripheral surrounding skin. In order to better understand variability in this context, three clinical experts generated ground truth segmentations on a subset of 100 images from the held-out test set. The Jaccard Index between the 3 pairs of clinicians was computed as a measure of agreement between experts. The resultant Jaccard index measurements were 0.743, 0.754, and 0.861, yielding an average of 0.786.

In summary, these results demonstrate that the segmentation network approach described here produced competitive segmentation performance to state-of-the-art, and showed agreement with the ground truth that was within the range of human experts, making it satisfactory for use in subsequent disease classification processing steps. The network was used as both a segmenter as well as a visual descriptor of lesion shape by saving the outputs of the most compressed fully-connected layer.

## *Classification*

Automated classification results are summarized in **Table 3**. Performance for systems utilizing each image context was assessed individually, including "Whole Image (WI)" (which considers the entire dermoscopic image), "Crop (CR)" (which considers only the bounding box tightly cropping the region segmented by the automatic segmentation), and "Crop GT (CRGT)" (which considers the bounding box tightly cropping the region segmented by the ground truth segmentation). Fusions of these individual systems are subsequently computed in "Part 3B: AVG/VOTE(WI, CRGT)," which combines the whole image context with the cropped region context from the ground truth segmentation, using average and voting fusions, respectively (and corresponds to Part 3B of the ISBI challenge), and "Part 3: AVG/VOTE(WI, CR)," which combines the whole image context with the cropped region context from the automatically generated segmentation, using average and voting fusions, respectively (and corresponds to Part 3 of the ISBI challenge tasks). The baseline prior state-of-the-art (winning challenge submission) for Parts 3 and 3B of the ISBI challenge are listed in the rows of the table labeled "Top Rank." In addition, two prior published methods, which we re-implemented and evaluated on this dataset, are listed in the bottom four rows of the table. These involve ensembles of low-level, unsupervised, and deep features without automated segmentation, as well as ensembles of low-level features alone.

While average precision results for Parts 3 and 3B demonstrate a 1.3% (0.645 vs. 0.637) and 4% (0.649 vs. 0.624) improvement over the prior state-of-the-art, AUC measurements exhibit a 4.2% (0.838 vs. 0.804) and 7.7% (0.843 vs 0.783) relative improvement, respectively. In addition, specificity measured at the clinically relevant sensitivity operating point of 95%, demonstrates a 43.6% (32.6% vs. 22.7%) and 194.4% (36.8% vs. 12.5%) relative improvement, respectively.

Curiously, the prior state-of-the-art showed improved average precision with automatically generated segmentation masks rather than relying on the ground truth segmentations (0.637 vs. 0.624). In this study, average precision performance of the system utilizing the ground truth segmentations showed a higher, but very similar, performance to that relying on automatically generated segmentation masks (0.649 vs. 0.645). These measurements may be the result of automated segmentation performance falling in the range of human performance and variability.

**Figure 5** shows the entire receiver operating characteristic (ROC) curve for both the proposed system and the prior state-of-the-art on both image classification tasks. Performance improvements are noted across multiple operating points, but particularly in areas corresponding to the highest sensitivity levels, which are clinically the most relevant system operating points.

The performance of the proposed system was also compared directly to the performance of expert dermatologists. On a random subset of 100 images in the test set (containing 50 melanomas and 50 non-melanomas), the average diagnostic performance of 8 dermatology experts, with an average 13.5 years of experience in dermoscopy (ranging from 6 to 27 years), was measured and reported [46]. Compared to the performance of these clinical experts on the subset of 100 images from the held-out test dataset, the proposed system using automatic segmentation produces a higher specificity performance level (62% vs. 59%) evaluated at an equivalent sensitivity operating point to the clinical experts (82%). In addition, evaluated at the learned threshold of 50% machine confidence, the system produced a higher accuracy than the average of the group of dermatologists (76% vs. 70.5%). When using ground truth segmentations, the system produced a specificity of 60% at the 82% sensitivity operating point, and an accuracy of 77% evaluated at the threshold of 50% machine confidence.

Next, we sought to assess the impact on system performance of using two ensemble component selection algorithms, as opposed to simple score averaging or voting of all components for ensemble fusion. While the initial selection of components to our system was informed by work done in the literature [22,23,32], we examined whether system performance can be further improved by selecting only those components of the system that lead to best performance on a 3-fold cross-validation within the training set. The two selection algorithms studied were greedy selection and forward model selection [47]. Greedy selection ranks each component based on its individual performance, combining them in order, until performance ceases to improve. Forward model selection runs in multiple iterations, where in each iteration, a search is performed to find the model that, when combined with the existing ensemble, improves recognition performance the most.

Our findings demonstrate that both selection algorithms, while modest in terms of the number of parameters learned, reduced performance on the held-out test set. Greedy model selection produced average precision for Parts 3 & 3B of 0.638 and 0.646, respectively. The intermediate steps and results of greedy model selection on the 3-fold cross-validation training data are shown in **Table 4**. Forward model selection produced 0.614 and 0.612, respectively. The intermediate steps and results of forward model selection on the 3-fold cross-validation training data are shown in **Table 5**. Between the two algorithms, forward model selection produced the best performance result on the 3-fold cross-validation dataset, but the worst performance on the held-out test dataset, suggesting that model selection routines might be overfitting the data. **Table 6** shows the average precision of each individual system component on the held-out test dataset, showing that each piece has learned a meaningful association with disease.

Finally, we quantified the contribution of the 1000-dimensions DRN concept detector vector to overall system performance by way of exclusion. If the component is individually removed from the ensemble of our system, performance decreases. Comparing the system using automatically generated segmentations, average precision drops from 0.645 to 0.632. Comparing the system using ground truth segmentations, the average precision drops from 0.649 to 0.633. These experiments confirm earlier reports that vectors of concept detectors from deep networks can contribute meaningful information to melanoma recognition in dermoscopy images [32].

## Discussion

There are a number of important insights that come from the results presented in this work. The first relates to the evaluation methodology of the classification tasks within ISBI challenge itself.

The organizers had decided to rank participants based on average precision, likely for its precedent in being commonly used to evaluate other public computer vision benchmarks, such as TRECVID [48]. While average precision is a useful retrieval metric for search based applications, it is clear from our results that it does not fully reflect the value of a system to the clinical workflow. As an example, our fully automated pipeline for Part 3 of the challenge was only 0.008 points in average precision above the prior state-of-the-art (a modest 1.3% improvement). However, inspecting the AUC metric revealed a difference of 0.034 (a 4.2% improvement). Focusing specifically on high sensitivity operating points of the ROC, where the system operation is clinically viable (so as to minimize the number of inadvertently missed disease states), specificity is increased by 9.9 points, representing over a 40% relative improvement. These comparisons, while clinically important and of pertinent use, are not reflected in the small change in average precision, suggesting that a different evaluation metric should be used to rank participants in subsequent challenges.

The second relates to the fundamental algorithm design as being an ensemble of a multitude of approaches. Since the high success rate of deep learning algorithms in the computer vision field, most technical research has focused on optimizing and building higher performing network structures. However, our results demonstrate that a traditional approach of generating ensembles across multiple techniques still holds value. For segmentation, we were able to create an ensemble of fully convolutional U-Nets to improve performance beyond that of the best performing single network itself. For classification, by combining a variety of both deep learning and classical computer vision techniques, our system demonstrated significant improvements over the state-of-the-art systems that used deep learning alone [35]. In the scope of this work, a naïve ensemble constituting simple prediction averaging among all components was found to outperform simple ensemble model selection methods such as greedy and forward model selection routines. This is likely an artifact of the small size of the dataset at this time, rather than the inherent utility of the selection methods. As the ISIC dataset grows, these ensemble learning experiments should be repeated, perhaps including more complex algorithms.

The third insight is that we have produced further evidence that complex non-linear data augmentation has rendered large neural networks better able to train on datasets with limited examples. The fully convolutional U-Net structure presented here contains over 543,888,390 parameters, and yet was trained from a dataset of only 900 examples (as a reference, VGG (Visual Geometry Group) Face [49], DeepFace [50], and FaceNet [51] used 2.6 million, 4.4 million, and 200 million faces, respectively, for training face recognition networks). When creating an ensemble of networks, the number of parameters further increases, roughly proportional to the number of networks. What is important to observe here is that the augmented data was not statically generated prior to training, but is generated dynamically during each mini-batch. In this manner, the size of the dataset effectively increases to infinity, as random perturbations are constantly introduced.

The fourth insight is that, surprisingly, deep residual networks trained on natural photographs from ImageNet contribute meaningful diagnostic information when the 1000 concept detector outputs are used as a descriptive image vector. This is consistent with past reports using similar methods from other networks [32]. In practice, it is common for human experts to describe patterns seen in lesions by analogies to other common objects in everyday life [32]. This was the initial intuition behind attempting to use these concept detectors as feature detectors in dermoscopic images. Our work is now the second published report to show positive results utilizing such a technique.

There are two limitations of the study. The first involves a lack of statistical significance of comparisons. Performance evaluations were carried out on a fixed dataset partition, which is necessary for a public challenge to maintain a held-out blind test dataset. In addition, software implementations of approaches used for comparison were not available to support multiple *n*-fold evaluations. The second limitation is that the diagnostic performance of clinical experts with dermoscopic images represents an approximation of performance in practice, where clinicians may benefit from the situational context of each lesion, including patient history, temporal evolution, comparison of the lesion to other lesions on the patient, and physical inspection.

## Conclusion and Future Work

In this paper, we have proposed a system for the segmentation and classification of melanoma from dermoscopic images of skin. The method was evaluated on the largest public benchmark for melanoma recognition available. New state-of-the-art performance is demonstrated, leading to an improvement in the area under receiver operating characteristic curve of 7.5% (0.843 vs. 0.783), in average precision of 4% (0.649 vs. 0.624), and in specificity measured at the clinically relevant 95% sensitivity operating point 2.9 times higher than the previous state-of-the-art (36.8% specificity compared to 12.5%). In addition, compared to the average disease recognition performance of 8 expert dermatologists, the proposed system produced a higher accuracy (76% vs. 70.5%) evaluated at the machine confidence threshold of 50%, and a higher specificity (62% vs. 59%) evaluated at the fixed sensitivity operating point of the clinicians (82%).

As the size of the ISIC dataset expands and additional dermoscopic pattern annotations become available, future work may consider learning a joint pattern-disease classification model, or building a semantic descriptor vector of dermoscopic patterns to be used in conjunction with other approaches for disease classification. In addition, the use of non-linear image warping as a data augmentation technique may be useful for classification. Other machine learning approaches may bring additional performance gains, such as residual convolutional layers for semantic segmentation, meta-learning or boosting for selection of network ensembles to perform segmentation, or use of these segmentation ensembles as more complex shape descriptors for disease classification. Finally, the use of additional situational contexts, such as patient history, patient metadata, temporal evolution, and comparison of the lesion to other lesions on the patient, should be studied, as all may further improve system performance.

## Acknowledgments

The authors would like to thank the larger community of the International Skin Imaging Collaboration (ISIC) for their effort in organizing the datasets used in this work, as well as engaging and insightful discussions in dermoscopy and dermatology. In addition, we would like to thank collaborators at Memorial Sloan-Kettering, including Ashfaq Marghoob, Michael Marchetti, and Stephen Dusza, for their expertise and support.

## Bios

**Noel C. F. Codella**  *IBM Research Division, Thomas J. Watson Research Center, Yorktown Heights, NY 10598 USA (nccodell@us.ibm.com).*  Dr. Codella is a Research Staff Member in the Multimedia Analytics Group in the Cognitive Computing organization at the IBM T. J. Watson Research Center. He received his B.S. degree in computer science from Columbia University in 2004, his M.Eng. degree in computer science from Cornell University in 2005, and his Ph.D. degree in physiology, biophysics, and systems biology from the Weill Cornell Medical College in 2010. Dr. Codella's thesis work included cardiac magnetic resonance imaging (MRI) free-breathing data acquisition, parallel MRI image reconstruction, and cardiac segmentation with functional analysis. He joined the IBM T. J. Watson Research Center in December 2010 as a Postdoctoral Researcher and later as Research Staff Member in October 2011. His research at IBM is focused on large-scale machine learning for visual analytics. He is the recipient of four IBM Invention Achievement Awards and two IBM Eminence and Excellence Awards, and he is co-recipient of an IBM Research Division Award. His work has generated over 40 peer-reviewed publications, 13 conference abstracts, and 14 patent filings, and his papers have been cited over 460 times.

**Quoc-Bao Nguyen**  *IBM Research Division, Thomas J. Watson Research Center, Yorktown Heights, NY 10598 USA (quocbao@us.ibm.com).* Mr. Nguyen is a Senior Software Engineer in the Cognitive Computing Department at the IBM T.J. Watson Research Center. He received a B.S. degree and an Engineering degree in computer sciences from the Conservatoire des Arts & Metiers, Paris (1987) and a post-graduate Diplôme d'études supérieures spécialisées (DESS) from Pierre & Marie Curie University, Paris (1988). He joined IBM in 1997 at the Thomas J. Watson Research Center where he has worked on rule-based diagnostic systems, dynamic scheduling of real-time tasks with social and temporal constraints, e-commerce activity-based profiling and user market segmentation, combinatorial optimization for reverse auctions, large-scale orchestration for distributed analytics systems, multi-modal search engines, and machine learning and deep learning on visual textual analysis.

**Sharath Pankanti**, *IBM Research Division, Thomas J. Watson Research Center, Yorktown Heights, NY 10598 USA (sharat@us.ibm.com).*  Dr. Pankanti is Principal Research Staff Member in the Cognitive Computing Department at the Thomas J. Watson Research Center.  He received his Ph.D. degree in computer science from the Michigan State University. His work contributed to the world's first large-scale biometric civilian fingerprint identification system in Peru and to an award-winning IBM surveillance offering that have been featured in the news media (ABC, Fox, CBS, and NBC), mentioned in popular TV media (*CSI: Miami*), and covered in social (good.is) media. He is a co-author of more than 150 peer-reviewed publications (more than 20,000 citations with h-index 48, per Google Scholar), which are published in many reputed venues, including *Scientific American*, *IEEE Computer*, *IEEE Spectrum*, *Comm. ACM*, and *Proc. IEEE.*  He is co-inventor of more than 100 inventions which has generated significant intellectual property revenue for IBM. Dr. Pankanti co-edited the first comprehensive book on biometrics, *Biometrics: Personal Identification* (Kluwer, 1999) and co-authored, *A Guide to Biometrics*" (Springer, 2004). He is Fellow of the IEEE, International Association of Pattern Recognition (IAPR), and Society of Photographic Instrumentation Engineers (SPIE). He has served the computer vision and pattern recognition community in various capacities over the last two decades and most recently volunteered as part of the IEEE Distinguished Visitor and ACM Distinguished Speaker programs.

**David A Gutman** *Department of Neurology, Emory Univesity School of Medicine, Atlanta, GA 30322 USA (dgutman@emory.edu)*. Dr. Gutman is an assistant professor in the Department of Neurology at Emory University in Atlanta. Dr. Gutman received his Bachelor's degree in chemistry from Penn State University, followed by an M.D./Ph.D (neuroscience) degree from Emory in 2005. Dr. Gutman then completed a psychiatry residency at Emory University in 2009. Dr. Gutman's thesis work focused on novel pharmaceutical treatments for depression and anxiety. Following residency, Dr. Gutman took a position as a research scientist in the Center for Comprehensive Informatics at Emory, and then transitioned into an assistant professorship in neurology and biomedical informatics. Dr. Gutman's work focuses primarily on imaging informatics, where he develops tools to store, annotate, and analyze large heterogeneous image sets. Dr. Gutman also built and maintains the Cancer Digital Slide Archive (http://cancer.digitalslidearchive.net), a public repository which hosts over 30,000 whole slide digital images.

**Brian Helba** *Kitware, Inc., Clifton Park, NY 12065 USA (brian.helba@kitware.com)*. Mr. Helba is an R&D Engineer at Kitware, Inc. He received a B.S. degree in biomedical engineering and computer science from Rensselaer Polytechnic Institute in 2010. Since 2011, Mr. Helba has designed and developed software and hardware technologies for quantitative computed tomography (CT) and positron emission tomography (PET) calibration, digital whole slide microscopy, optical coherence tomography (OCT) visualization, and medical image data management and analysis. His work has appeared in multiple peer-reviewed publications, conference proceedings, and patent filings. He is the lead developer of the International Skin Imaging Collaboration (ISIC) Archive and a core maintainer of multiple widely used open source software toolkits.

**Allan C. Halpern** *Memorial Sloan Kettering Cancer Center. New York, NY 10065 USA (halperna@mskcc.org)*. Dr. Halpern is Chief of the Dermatology Service and co-leader of the Melanoma Disease Management Team at Memorial Sloan Kettering Cancer Center. Dr. Halpern has contributed to the understanding of melanoma prognosis and risk stratification as well as the natural history of melanocytic tumor progression in humans. Dr. Halpern is an internationally recognized expert in imaging technologies relevant to clinical care and epidemiologic studies. He has developed an optical imaging laboratory at Memorial Sloan Kettering Cancer Center and is an internationally recognized leader in the areas of total body imaging, dermoscopy, and in-vivo confocal microscopy. He is president of the International Society for Digital Imaging of the Skin, a member of the executive board of the International Dermoscopy Society, and he initiated and leads the International Skin Imaging Collaboration.

**John R. Smith** *IBM Research Division, Thomas J. Watson Research Center, Yorktown Heights, NY 10598 USA (jsmith@us.ibm.com)*. Dr. Smith is an IBM Fellow and Manager of Multimedia and Vision team at IBM T. J. Watson Research Center. He leads IBM's research related to IBM Multimedia Analysis and Retrieval System (IMARS), Intelligent Video Analytics (IVA), and IBM Watson Developer Cloud Visual Recognition, among other industry projects related to multimedia and computer vision. Dr. Smith was Editor-in-Chief of *IEEE Multimedia* from 2010 to 2014 and served as co-General Chair of ACM International Conference on Multimedia Retrieval (ICMR-2016) in New York City. Dr. Smith is a Fellow of IEEE.

Figures and Tables

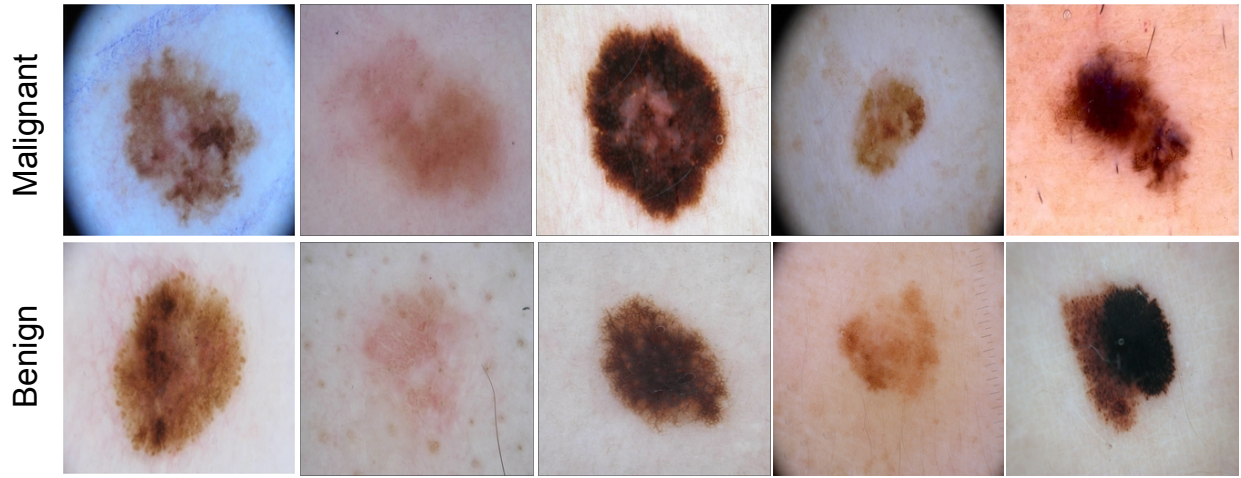

**Figure 1:** Example dermoscopic images from the ISBI 2016 Challenge "Skin Lesion Analysis Toward Melanoma Detection." Dermoscopy provides high-resolution magnified images of skin without interference from surface skin reflection, allowing visualization of finely detailed dermatological structures. Even with such professional grade images, distinction between disease and non-diseased lesions is a difficult task. *Top row*: malignant melanomas. *Bottom row*: benign nevi.

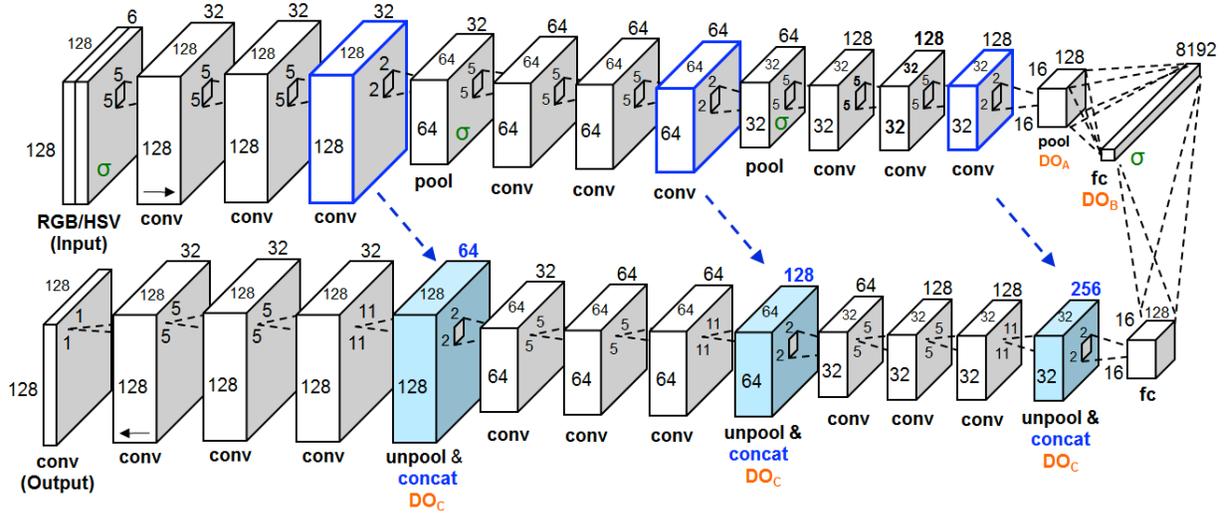

**Figure 2**: Visual depiction of the implemented fully-convolutional U-Net structure with joint RGB and HSV channel inputs, with optimized parameters (numbers) shown. Convolutional and deconvolutional layers are marked as "conv" layers, with dotted lines and small grey boxes depicting each operation on the previous layer. Pooling layers are marked as "pool," whereas unpooling layers are marked with "unpool." The sizes of the convolution and pooling operations are shown nearby the dotted lines and small boxes representing them. Layers with outputs that are summed with zero-mean Gaussian noise are marked with green sigma. Layers with outputs that undergo dropout are marked with "DO" (with subscript referring to the parameter exposed to user to control degree of dropout). Layers undergoing concatenation with skip connections are shaded blue and marked as "concat" layers. The sources of the skip connections for the concatenation operations are outlined in solid blue. The number of convolution kernels in each layer is shown at the top of the layer, and the spatial resolution of each layer is written near each dimension edge.

**Table 1**: Segmentation results, in terms of Jaccard Index and pixel-wise accuracy, of four variants of fully-convolutional U-Net approaches, followed by state-of-art on the ISBI dataset, and of inter-observer human expert agreement.

| Method | Jaccard | Accuracy |
|---|---|---|
| *Optimized Single* | 0.836 | 94.9% |
| *Default Augmentation* | 0.828 | 94.7% |
| *No Noise or Dropout* | 0.812 | 94.1% |
| *Ensemble of 10 U-Nets* | 0.841 | 95.1% |
| *State-of-art* | 0.843 | 95.3% |
| *Human Expert Average Agreement* | 0.786 | 90.9% |

**Table 2**: Fully-convolutional U-Net training parameters used in ensemble segmentation approach. (FC Dims.: fully connected layer dimensionality; DO: dropout (refer to Figure 2 for subscript interpretation); Mo.: momentum; Ep.: epochs.)

| Input Size | Kernel Size | Pool Size | # Conv Filters | FC Dim | $DO_A$ | $DO_B$ | $DO_C$ | Noise | Learn Rate | Mo. | Max Ep. | Valid Loss |
|---|---|---|---|---|---|---|---|---|---|---|---|---|
| 128 | 5 | 2 | 32 | 8192 | 0.5 | 0.5 | 0.5 | 0.025 | 0.01 | 0.95 | 2000 | 0.096 |
| 128 | 5 | 2 | 32 | 4096 | 0.5 | 0.5 | 0.5 | 0.025 | 0.01 | 0.95 | 2000 | 0.102 |
| 128 | 5 | 2 | 32 | 2048 | 0.5 | 0.5 | 0.5 | 0.025 | 0.01 | 0.95 | 2000 | 0.107 |
| 128 | 5 | 2 | 32 | 1024 | 0.5 | 0.5 | 0.5 | 0.025 | 0.01 | 0.95 | 2000 | 0.111 |
| 128 | 5 | 2 | 32 | 512 | 0.5 | 0.5 | 0.5 | 0.025 | 0.01 | 0.95 | 2000 | 0.106 |
| 128 | 5 | 2 | 32 | 256 | 0.5 | 0.5 | 0.5 | 0.025 | 0.01 | 0.95 | 2000 | 0.101 |
| 128 | 3 | 2 | 16 | 1024 | 0.5 | 0.5 | 0.5 | 0.025 | 0.01 | 0.95 | 2000 | 0.119 |
| 128 | 5 | 2 | 32 | 8192 | 0.5 | 0.75 | 0.5 | 0.025 | 0.01 | 0.95 | 2000 | 0.126 |
| 128 | 5 | 2 | 32 | 8192 | 0.25 | 0.5 | 0.25 | 0.025 | 0.01 | 0.95 | 2000 | 0.104 |
| 64 | 5 | 2 | 32 | 8192 | 0.5 | 0.5 | 0.5 | 0.025 | 0.01 | 0.95 | 2000 | 0.095 |

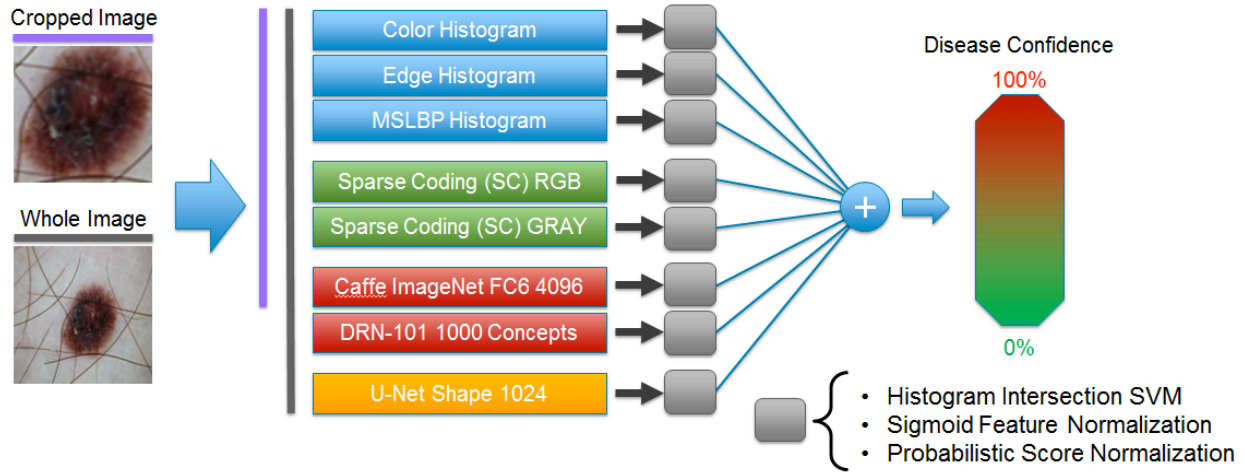

**Figure 3**: Visualization of classification framework. Various features—from rule-based feature extractors, unsupervised learning systems, or deep neural nets—are extracted across two scales: an area cropped around the lesion, and the entire dermoscopic image. Non-linear SVMs (gray boxes) are trained over these features, probabilistically normalized, and averaged in an ensemble, to produce a final disease risk score between 0.0 – 1.0 (0% - 100%). The Caffe and 101 layer Deep Residual Network (DRN-101) are trained from ImageNet data. (MSLBP: multiscale local binary patterns.)

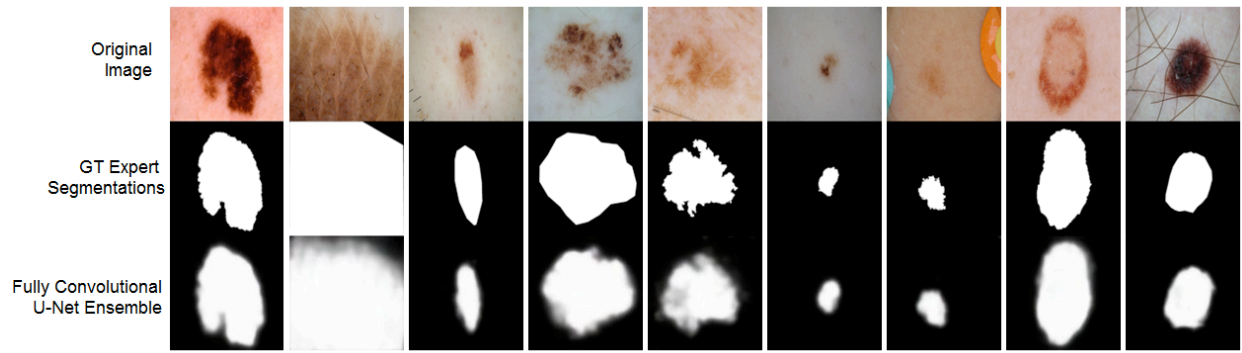

**Figure 4**: Example results of the fully-convolutional U-Net Ensemble based segmentation network. Often, lesions are complex structures with ill-defined boundaries, making the segmentation task non-trivial. The network performs well even under these difficult circumstances. *Top row*: original image. *Middle Row*: ground truth segmentations. White pixels (intensity 255) depict areas inside the lesion, and black pixels (intensity 0) depict areas outside the lesion (background normal skin). *Bottom row:* output of the U-Net Ensemble. The network produces confidence scores that each pixel belongs to a lesion, with most confident being white, least confident being black, and varying degrees of confidence between these two values. The spectrum of confidence levels leads to a blurry appearance in the output.

**Table 3**: Classification performance results, with comparison to the state-of-the-art. (AP: average precision; ACC: accuracy; SENS: sensitivity; SPEC: specificity; SP95: specificity evaluated at 95% sensitivity; AUC: area under receiver operator curve; AVG(): average score fusion; VOTE(): voting fusion.)

| *Method* | *AP* | *ACC* | *SENS* | *SPEC* | *SP95* | *AUC* |
| --- | --- | --- | --- | --- | --- | --- |
| *Whole Image (WI)* | 0.596 | 0.755 | 0.627 | 0.796 | 0.319 | 0.808 |
| *Crop (CR)* | 0.618 | 0.81 | 0.72 | 0.832 | 0.312 | 0.819 |
| *Crop GT (CRGT)* | 0.629 | 0.781 | 0.707 | 0.799 | 0.382 | 0.827 |
| *Part 3B: VOTE(WI, CRGT)* | 0.588 | 0.834 | 0.533 | 0.9079 | 0.359 | 0.829 |
| *Part 3: VOTE(WI, CR)* | 0.602 | 0.834 | 0.52 | 0.9112 | 0.306 | 0.828 |
| *Part 3B: AVG(WI, CRGT)* | 0.649 | 0.807 | 0.693 | 0.836 | 0.368 | 0.843 |
| *Part 3: AVG(WI, CR)* | 0.645 | 0.805 | 0.693 | 0.832 | 0.326 | 0.838 |
| *Part 3B: Top Rank [35]* | 0.624 | 0.855 | 0.547 | 0.931 | 0.125 | 0.783 |
| *Part 3: Top Rank [35]* | 0.637 | 0.855 | 0.507 | 0.941 | 0.227 | 0.804 |
| *Part 3B: Codella et. al. [32]* | 0.591 | 0.836 | 0.253 | 0.98 | 0.312 | 0.815 |
| *Part 3: Codella et. al. [32]* | 0.589 | 0.77 | 0.72 | 0.723 | 0.174 | 0.815 |
| *Part 3B: LL Ensemble [32]* | 0.532 | 0.726 | 0.693 | 0.734 | 0.151 | 0.603 |
| *Part 3: LL Ensemble [32]* | 0.506 | 0.752 | 0.64 | 0.78 | 0.112 | 0.643 |

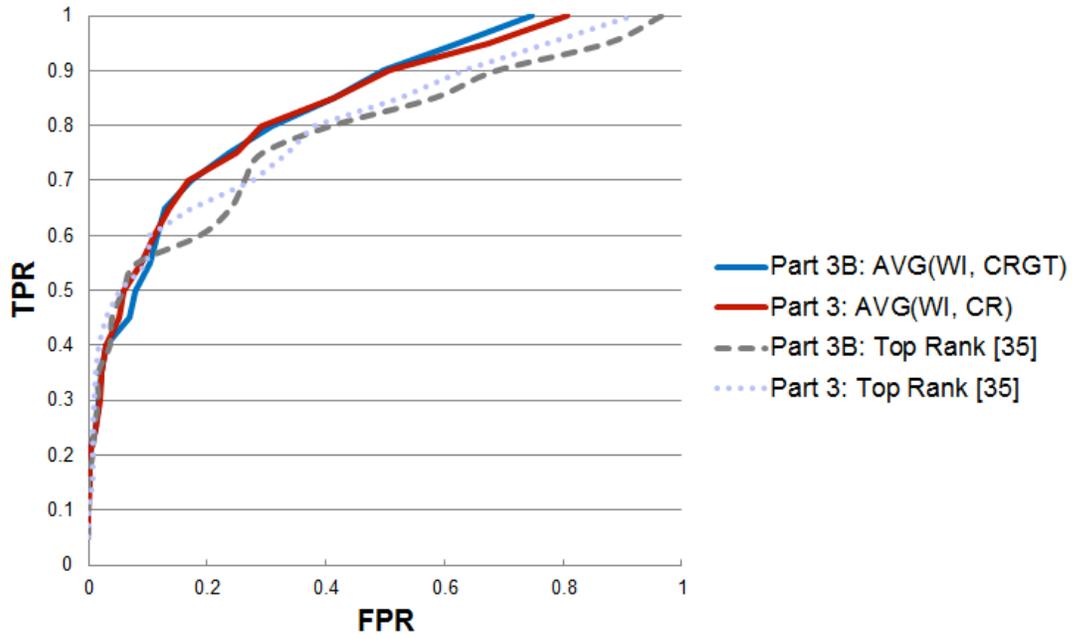

**Figure 5**: Receiver operating characteristic (ROC) curve plots for the proposed scheme and the prior state-of-the-art on both image classification tasks (3 & 3B) of the ISBI 2016 Challenge on Skin Lesion Analysis Toward Melanoma Detection. (TPR: true positive rate; FPR: false positive rate.)

**Table 4:** Iterative steps of greedy model selection using 3-fold cross-validation scores on the training dataset. Features are sorted according to average precision (AP), and subsequently combined in order of performance. The cumulative AP shows the performance of the ensemble when that model and all better performing models have been combined. The final ensemble model cumulative AP is highlighted in bold. (CRGT: cropped ground truth context; WI: whole image context; SC RGB: color sparse codes; SC GRAY: grayscale sparse codes; MSLBP: multiscale local binary patterns; DRN: deep residual network; FC6: 6$^{th}$ fully connected layer of Caffe model.)

| *Context* | *Feature* | *Individual AP* | *Cumulative AP* |
| --- | --- | --- | --- |
| *CRGT* | *Sparse Coding RGB* | 0.472 | 0.472 |
| *CRGT* | *Caffe FC6* | 0.46 | 0.504 |
| *WI* | *Caffe FC6* | 0.453 | 0.532 |
| *CRGT* | *SC GRAY* | 0.389 | 0.538 |
| *WI* | *DRN 1K Concepts* | 0.361 | 0.535 |
| *WI* | *Sparse Coding RGB* | 0.36 | 0.53 |
| *WI* | *MSLBP* | 0.359 | 0.545 |
| *CRGT* | *Color Histogram* | 0.349 | 0.558 |
| *CRGT* | *MSLBP* | 0.347 | 0.56 |
| *WI* | *U-Net Shape 1024* | 0.345 | **0.567** |
| *WI* | *SC GRAY* | 0.34 | 0.564 |
| *WI* | *Color Histogram* | 0.3 | 0.562 |
| *WI* | *Edge Histogram* | 0.298 | 0.567 |
| *CRGT* | *Edge Histogram* | 0.239 | 0.565 |

**Table 5**: Iterative steps of forward model selection process using 3-fold cross-validation on the training dataset. In the first iteration, the single model with maximum performance was chosen (highlighted in bold). In the subsequent iterations, remaining models were searched—the model that maximizes performance the most is selected for inclusion in the ensemble. The process stops when performance no longer improves (iteration 8). (CRGT: cropped ground truth context; WI: whole image context; SC RGB: color sparse codes; SC GRAY: grayscale sparse codes; MSLBP: multiscale local binary patterns; DRN: deep residual network; FC6: 6[th] fully connected layer of Caffe model.)

| Context | Feature | Iteration: 1 AP | 2 AP | 3 AP | 4 AP | 5 AP | 6 AP | 7 AP | 8 AP |
|---|---|---|---|---|---|---|---|---|---|
| **CRGT** | **SC RGB** | **0.472** | | | | | | | |
| *CRGT* | *Caffe FC6* | 0.46 | 0.504 | 0.532 | **0.562** | | | | |
| *WI* | *Caffe FC6* | 0.453 | **0.531** | | | | | | |
| CRGT | SC GRAY | 0.389 | 0.489 | 0.532 | 0.555 | 0.56 | 0.571 | 0.576 | 0.576 |
| WI | DRN 1K | 0.361 | 0.462 | 0.516 | 0.533 | 0.545 | 0.558 | 0.565 | 0.57 |
| WI | SC RGB | 0.36 | 0.475 | 0.524 | 0.543 | 0.552 | 0.561 | 0.566 | 0.569 |
| *WI* | *MSLBP* | 0.359 | 0.519 | **0.559** | | | | | |
| CRGT | Color Hist. | 0.349 | 0.488 | 0.537 | 0.559 | 0.567 | 0.574 | 0.576 | 0.574 |
| *CRGT* | *MSLBP* | 0.347 | 0.484 | 0.534 | 0.549 | 0.565 | 0.572 | **0.579** | |
| *WI* | *U-Net 1024* | 0.345 | 0.465 | 0.531 | 0.555 | 0.568 | **0.576** | | |
| WI | SC GRAY | 0.34 | 0.482 | 0.531 | 0.548 | 0.557 | 0.563 | 0.569 | 0.574 |
| WI | Color Hist. | 0.3 | 0.469 | 0.522 | 0.548 | 0.556 | 0.564 | 0.565 | 0.57 |
| *WI* | *Edge Hist.* | 0.298 | 0.474 | 0.539 | 0.559 | **0.572** | | | |
| CRGT | Edge Hist. | 0.239 | 0.427 | 0.501 | 0.534 | 0.552 | 0.554 | 0.562 | 0.564 |

**Table 6:** Performance of individual machine learning components on the 379 held-out test set images. (CRGT: Cropped Ground Truth Context; WI: Whole Image Context; SC RGB: Color Sparse Codes; SC GRAY: Grayscale Sparse Codes; MSLBP: Multiscale Local Binary Patterns; DRN: Deep Residual Network; FC6: 6$^{th}$ fully connected layer of Caffe model.)

| Context | Feature | AP | ACC | SENS | SPEC | AUC |
|---|---|---|---|---|---|---|
| CRGT | Color Histogram | 0.36 | 0.789 | 0.213 | 0.9309 | 0.626 |
| | Caffe FC6 | 0.504 | 0.734 | 0.707 | 0.74 | 0.787 |
| | SC GRAY | 0.457 | 0.694 | 0.707 | 0.691 | 0.762 |
| | SC RGB | 0.435 | 0.702 | 0.64 | 0.717 | 0.736 |
| | Edge Histogram | 0.265 | 0.665 | 0.4 | 0.73 | 0.571 |
| | MSLBP | 0.479 | 0.694 | 0.6 | 0.717 | 0.716 |
| | | | | | | |
| WI | Color Histogram | 0.333 | 0.776 | 0.267 | 0.9013 | 0.615 |
| | Caffe FC6 | 0.488 | 0.723 | 0.693 | 0.73 | 0.764 |
| | SC GRAY | 0.449 | 0.678 | 0.627 | 0.691 | 0.736 |
| | SC RGB | 0.447 | 0.699 | 0.6 | 0.724 | 0.75 |
| | DRN 1K Concepts | 0.466 | 0.726 | 0.507 | 0.78 | 0.749 |
| | Edge Histogram | 0.287 | 0.686 | 0.48 | 0.737 | 0.593 |
| | MSLBP | 0.416 | 0.694 | 0.587 | 0.72 | 0.723 |
| | U-Net 1024 | 0.375 | 0.702 | 0.573 | 0.734 | 0.715 |